\documentclass[11pt,a4paper]{article}
\usepackage[hyperref]{acl2018}

\usepackage{times}
\usepackage{url}
\usepackage{latexsym}
\usepackage{color,soul}
\usepackage{amsmath}
\usepackage{graphicx}
\usepackage{array}
\usepackage{float}
\usepackage{algorithm}
\usepackage[noend]{algpseudocode}
\usepackage{pgfplotstable}
\usepackage{pgfplots}
\pgfplotsset{compat=1.14}
\usepackage{multirow}
\usepackage{pbox}
\usepackage{amssymb}
\usepackage{caption}
\usepackage{mdframed}
\usepackage{hhline}
\usepackage{boxedminipage}
\usepackage{tikz-qtree,tikz-qtree-compat}
\usepackage{subcaption}
\usepackage{xcolor}
\usepackage{tikz}
\usepackage{multirow}
\usepackage{cleveref}
\usepackage{blindtext}
\usepackage[absolute]{textpos}

\input{insbox}

\newcommand\task{{\sc S}plit-and-{\sc R}ephrase}

\newcommand\websplit{\textsc{WebSplit}}

\newcommand\seqseq{\textsc{Seq2Seq}}
\newcommand\mseqseq{\textsc{MultiSeq2Seq}}
\newcommand\hybridsimpl{\textsc{HybridSimpl}}
\newcommand\splitseqseq{\textsc{Split-Seq2Seq}}
\newcommand\splitmseqseq{\textsc{Split-MultiSeq2Seq}}
\newcommand\bleu{\textsc{BLEU}}

\definecolor{darkgreen}{rgb}{0.0, 0.5, 0.0}

\newcolumntype{P}[1]{>{\centering\arraybackslash}p{#1}}

\aclfinalcopy % Uncomment this line for the final submission
\def\aclpaperid{1221} %  Enter the acl Paper ID here

\usepackage{fancyhdr}
\usepackage{geometry}
\usepackage{layout}
\title{Split and Rephrase: Better Evaluation and a Stronger Baseline}

\author{Roee Aharoni \& Yoav Goldberg \\
Computer Science Department\\
Bar-Ilan University\\
Ramat-Gan, Israel \\
\texttt{\{roee.aharoni,yoav.goldberg\}@gmail.com} \\
}

\date{}

\begin{document}

\maketitle
\global\csname @topnum\endcsname 0
\global\csname @botnum\endcsname 0
\begin{abstract}
Splitting and rephrasing a complex sentence into several shorter sentences that convey the same meaning is a challenging problem in NLP. We show that while vanilla seq2seq models can reach high scores on the proposed benchmark \cite{shashi2017}, they suffer from memorization of the training set which contains more than 89\% of the unique simple sentences from the validation and test sets. To aid this, we present a new train-development-test data split and neural models augmented with a copy-mechanism, outperforming the best reported baseline by 8.68 BLEU and fostering further progress on the task.
\end{abstract}

% "accepted..." header
\thispagestyle{fancy}

\section{Introduction}
\label{sec:intro}

% what is it and why is it important
% sprp is important for many tasks
Processing long, complex sentences is challenging. This is true either for humans in various circumstances \cite{inui2003text, watanabe2009facilita,de2010text} or in NLP tasks like parsing \cite{tomita1986efficient, mcdonald2011analyzing, jelinek2014improvements} and machine translation \cite{chandrasekar1996motivations,pougetabadie-EtAl:2014:SSST-8, koehn2017six}. An automatic system capable of breaking a complex sentence into several simple sentences that convey the same meaning is very appealing. 

% what was done previously
A recent work by \newcite{shashi2017} introduced a dataset, evaluation method and baseline systems for the task, naming it ``\task{}''. The dataset includes 1,066,115 instances mapping a single complex sentence to a sequence of sentences that express the same meaning, together with RDF triples that describe their semantics.
% why is it not solved
They considered two system setups: a text-to-text setup that does not use the accompanying RDF information, and a semantics-augmented setup that does. They report a BLEU score of 48.9 for their best text-to-text system, and of 78.7 for the best RDF-aware one. We focus on the text-to-text setup, which we find to be more challenging and more natural.

% our approach
We begin with vanilla \seqseq{} models with attention \cite{Bahdanau2014NeuralMT} and reach an accuracy of 77.5 BLEU, substantially outperforming the text-to-text baseline of \newcite{shashi2017} and approaching their best RDF-aware method. However, manual inspection reveal many cases of unwanted behaviors in the resulting outputs: (1) many resulting sentences are \emph{unsupported} by the input: they contain correct facts about relevant entities, but these facts were not mentioned in the input sentence; (2) some facts are \emph{repeated}---the same fact is mentioned in multiple output sentences; and (3) some facts are \emph{missing}---mentioned in the input but omitted in the output. 

% analysis
The model learned to \emph{memorize entity-fact pairs} instead of learning to split and rephrase. Indeed, feeding the model with examples containing entities alone without any facts about them causes it to output perfectly phrased but unsupported facts (Table \ref{tab:issues}). Digging further, we find that 99\% of the simple sentences (more than 89\% of the unique ones) in the validation and test sets also appear in the training set, which---coupled with the good memorization capabilities of \seqseq{} models and the relatively small number of distinct simple sentences---helps to explain the high \bleu{} score.

% improvements
To aid further research on the task, we propose a more challenging split of the data. We also establish a stronger baseline by extending the \seqseq{} approach with a copy mechanism, which was shown to be helpful in similar tasks \cite{gu2016incorporating,merity2016pointer,abc2017}. On the original split, our models outperform the best baseline of \newcite{shashi2017} by up to 8.68 BLEU, without using the RDF triples. On the new split, the vanilla \textsc{Seq2Seq} models break completely, while the copy-augmented models perform better. In parallel to our work, an updated version of the dataset was released (v1.0), which is larger and features a train/test split protocol which is similar to our proposal. We report results on this dataset as well.
% We also report a first result on the recently released v1.0 of the dataset which is larger and does not suffer from the sentence overlap issues. 
The code and data to reproduce our results are available on Github.\footnote{\url{https://github.com/biu-nlp/sprp-acl2018}} We encourage future work on the split-and-rephrase task to use our new data split or the v1.0 split instead of the original one.

\section{Preliminary Experiments}
\paragraph{Task Definition} In the split-and-rephrase task we are given a complex sentence $C$, and need to produce a sequence of simple sentences $T_1,...,T_n$, $n\geq2$, such that the output sentences convey all and only the information in $C$. As additional supervision, the split-and-rephrase dataset associates each sentence with a set of RDF triples that describe the information in the sentence. Note that the number of simple sentences to generate is not given as part of the input.

% websplit statistics
\begin{table}[tbp]
\centering
{\footnotesize
\centering
\begin{tabular}{l|l|l}
                         & count & unique \\\hline
RDF entities             & 32,186    & 925     \\
RDF relations            & 16,093	 & 172     \\
complex sentences        & 1,066,115 & 5,544     \\
simple sentences         & 5,320,716 & 9,552     \\
\hline
train complex sentences  & 886,857   & 4,438   \\
train simple sentences   & 4,451,959 & 8,840   \\
dev complex sentences    & 97,950    & 554     \\ 
dev simple sentences     & 475,337   & 3,765   \\ 
test complex sentences   & 81,308    & 554     \\
test simple sentences    & 393,420   & 4,015   \\\hline 
\% dev simple in train   & 99.69\%   & 90.9\%  \\
\% test simple in train  & 99.09\%   & 89.8\%  \\
\% dev vocab in train    &\multicolumn{2}{l}{ 97.24\% } \\
\% test vocab in train    &\multicolumn{2}{l}{ 96.35\% } \\

\end{tabular}
}
\caption{Statistics for the \websplit{} dataset.}
\label{tab:stats}
\end{table}

% preliminary results
% reference statistics from preprocess.py/train_dev_test_statistics() 
\begin{table}[htbp!]{
\centering
\footnotesize
\begin{tabular}{l | r r r }
Model & BLEU & \#S/C & \#T/S \\\hline
\textsc{Source} & 55.67 & 1.0 & 21.11  \\
\textsc{Reference} & -- & 2.52 & 10.93  \\\hline
\newcite{shashi2017} & \multicolumn{3}{l}{}  \\
\hybridsimpl & 39.97 & 1.26 & 17.55 \\
\seqseq & 48.92 & 2.51 & 10.32  \\
\mseqseq* & 42.18 & 2.53 & 10.69  \\
\splitmseqseq* & 77.27 & 2.84 & 11.63 \\ 
\splitseqseq* & 78.77 & 2.84 & 9.28 \\
\hline
This work & \multicolumn{3}{l}{}  \\
\textsc{Seq2Seq128} & 76.56 & 2.53 & 10.53  \\ %onmt
\textsc{Seq2Seq256} & 77.48 & 2.57 & 10.56  \\ %onmt
\textsc{Seq2Seq512} & 75.92 & 2.59 & 10.59  \\ %onmt
% \textsc{Seq2Seq512} & 80.39 & 2.9 & 9.9  \\ %marian
% \textsc{Seq2Seq1024} & 82.97 & 2.9 & 10.02 \\ %marian
\end{tabular}}
\caption{BLEU scores, simple sentences per complex sentence (\#S/C) and tokens per simple sentence (\#T/S), as computed over the test set. \textsc{Source} are the complex sentences and \textsc{Reference} are the reference rephrasings from the test set. Models marked with * use the semantic RDF triples.}\label{tab:results}
\vspace{-0.5cm}
\end{table}

\paragraph{Experimental Details}\label{sec:experimental_details}
We focus on the task of splitting a complex sentence into several simple ones \emph{without} access to the corresponding RDF triples in either train or test time. For evaluation we follow \newcite{shashi2017} and compute the averaged individual multi-reference BLEU score for each prediction.\footnote{Note that this differs from "normal" multi-reference BLEU (as implemented in \path{multi-bleu.pl}) since the number of references differs among the instances in the test-set.} We split each prediction to sentences\footnote{Using \textsc{NLTK} v3.2.5 \url{https://www.nltk.org/}} and report the average number of simple sentences in each prediction, and the average number of tokens for each simple sentence.
% \paragraph{Models}
% \label{sec:preliminary_models}
We train vanilla sequence-to-sequence models with attention \cite{Bahdanau2014NeuralMT} as implemented in the \textsc{Opennmt-py} toolkit \cite{2017opennmt}.\footnote{\url{https://github.com/OpenNMT/OpenNMT-py} commit d4ab35a} Our models only differ in the LSTM cell size (128, 256 and 512, respectively). See the supplementary material for training details and hyperparameters. 
% Adam with $\alpha=10^{-8},\beta_{1}=0.9,\beta_{2}=0.999$.
% \paragraph{Baselines}
% baselines
We compare our models to the baselines proposed in \newcite{shashi2017}. \hybridsimpl{} and \seqseq{} are text-to-text models, while the other reported baselines additionally use the RDF information.

% examples table
\begin{table*}[tbp]
\centering
{\footnotesize
\begin{tabular}{p{5.5cm}|p{9.5cm}}

% \hline
Input & Prediction \\
\hline
A Fortress of Grey Ice {\color{blue}with ISBM 0-7653-0633-6} has 672 pages .&\begin{tabular}[t]{p{9cm}} \vspace{-0.2cm}{\color{red}J.V. Jones authored A Fortress of Grey Ice .}\\ A Fortress of Grey Ice has 672 pages .
\end{tabular} \\
\hline
The address , 11 Diagonal Street is located in South Africa where the leader is Cyril Ramaphosa {\color{blue}and some Asian South Africans live} . &
\begin{tabular}[t]{p{9.3cm}}
The address , 11 Diagonal Street is located in South Africa . \\
The leader of South Africa is called Cyril Ramaphosa . \\
\vspace{-0.26cm}{\color{darkgreen}The leader of South Africa is called Cyril Ramaphosa .}\\
\vspace{-0.26cm}{\color{darkgreen}The leader of South Africa is called Cyril Ramaphosa .}\\
\end{tabular} \\ 
\hhline{=|=}
Alan Shepard Alan Shepard Alan Shepard
 & 
\begin{tabular}[t]{p{9.3cm}}
\vspace{-0.26cm}{\color{red}Alan Shepard is dead .} \\\vspace{-0.27cm}{\color{red}Alan Shepard was a test pilot .}
\end{tabular} \\ \hline
AFC Ajax AFC Ajax AFC Ajax
 &\begin{tabular}[t]{p{9cm}}
\vspace{-0.26cm}{\color{red}AFC Ajax 's manager is Jong Ajax .}\\\vspace{-0.26cm}{\color{red}AFC Ajax N.V. own Sportpark De Toekomst .}\end{tabular}\\
% \hline
\end{tabular}
}%end footnotesize
\vspace{-5px}
\caption{Predictions from a vanilla \seqseq{} model, illustrating {\color{red}unsupported facts}, {\color{blue}missing facts} and {\color{darkgreen}repeated facts}. The last two rows show inputs we composed to demonstrate that the models memorize entity-fact pairs.}  
\label{tab:issues}
\vspace{-15px}
\end{table*}

\paragraph{Results}
As shown in Table \ref{tab:results}, our 3 models obtain higher BLEU scores then the \seqseq{} baseline, with up to 28.35 BLEU improvement, despite being single-layer models vs. the 3-layer models used in \newcite{shashi2017}. A possible explanation for this discrepancy is the \seqseq{} baseline using a dropout rate of 0.8, while we use 0.3 and only apply it on the LSTM outputs. Our results are also better than the \mseqseq{} and \splitmseqseq{} models, which use explicit RDF information. We also present the macro-average\footnote{Since the number of references varies greatly from one complex sentence to another, (min: 1, max: 76,283, median: 16) we avoid bias towards the complex sentences with many references by performing macro average, i.e. we first average the number of simple sentences in each reference among the references of a specific complex sentence, and then average these numbers.} number of simple sentences per complex sentence in the reference rephrasings (\textsc{Reference}) showing that the \splitmseqseq{} and \splitseqseq{} baselines may suffer from over-splitting since the reference splits include 2.52 simple sentences on average, while the mentioned models produced 2.84 sentences.
\vspace{-0.6cm}
% vanilla attn weights figure
{
\InsertBoxC{\begin{tabular}{c}\\
\includegraphics[scale=0.34]{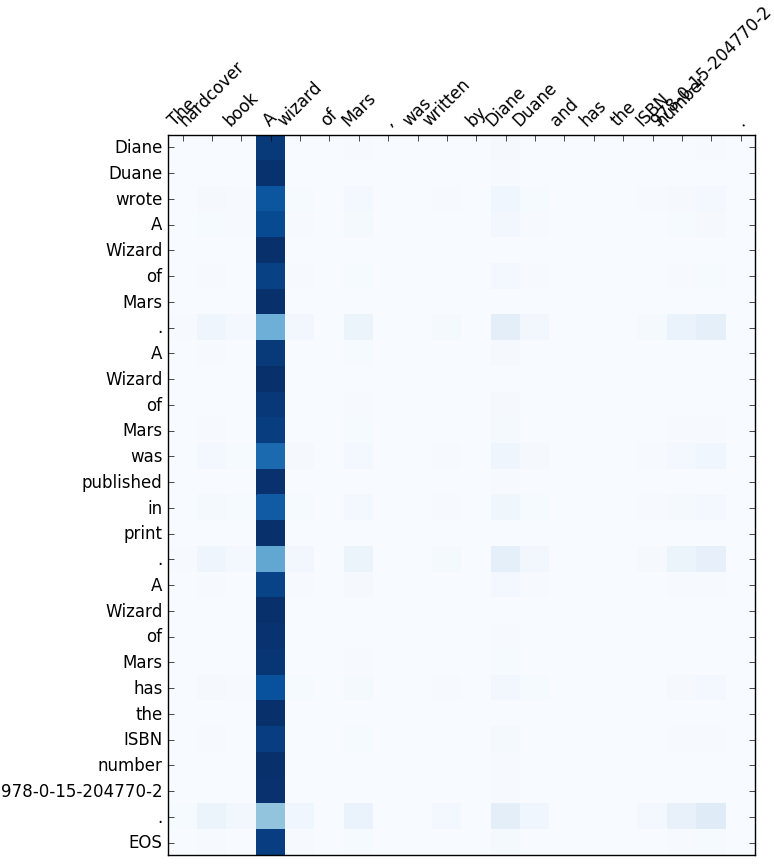}\\
\captionsetup{type=figure}
\parbox{\linewidth}{\captionof{figure}{\textsc{Seq2Seq512}'s attention weights. Horizontal: input. Vertical: predictions.} 
\vspace{-0.3cm}
\label{fig:vanilla_attn}}\end{tabular}}
}
\paragraph{Analysis}\label{sec:analysis}
We begin analyzing the results by manually inspecting the model's predictions on the validation set. This reveals three common kinds of mistakes as demonstrated in Table \ref{tab:issues}: unsupported facts, repetitions, and missing facts. All the unsupported facts seem to be related to entities mentioned in the source sentence.
% \paragraph{Skewed Attention} 
Inspecting the attention weights (Figure \ref{fig:vanilla_attn}) reveals a worrying trend: throughout the prediction, the model focuses heavily on the first word in of the first entity (``A wizard of Mars'') while paying little attention to other cues like ``hardcover'', ``Diane'' and ``the ISBN number''.
% \paragraph{Memorization}
This explains the abundance of ``hallucinated'' unsupported facts: rather than learning to split and rephrase, the model learned to identify entities, and spit out a list of facts it had memorized about them. 
To validate this assumption, we count the number of predicted sentences which appeared as-is in the training data. We find that 1645 out of the 1693 (97.16\%) predicted sentences appear verbatim in the training set. Table \ref{tab:stats} gives more detailed statistics on the \websplit{} dataset.

To further illustrate the model's recognize-and-spit strategy, we compose inputs containing an entity string which is duplicated three times, as shown in the bottom two rows of Table \ref{tab:issues}. As expected, the model predicted perfectly phrased and correct facts about the given entities, although these facts are clearly not supported by the input.

\section{New Data-split}\label{sec:split}
The original data-split is not suitable for measuring generalization, as it is susceptible to ``cheating'' by fact memorization. We construct a new train-development-test split to better reflect our expected behavior from a split-and-rephrase model.
We split the data into train, development and test sets by randomly dividing the 5,554 distinct complex sentences across the sets, while using the provided RDF information to ensure that:
\begin{enumerate}
\item Every possible RDF relation (e.g., \textsc{BornIn}, \textsc{LocatedIn}) is represented in the training set (and may appear also in the other sets).
\item Every RDF triplet (a complete fact) is represented only in one of the splits.
\end{enumerate}

While the set of complex sentences is still divided roughly to 80\%/10\%/10\% as in the original split, now there are nearly no simple sentences in the development and test sets that appear verbatim in the train-set. Yet, every relation appearing in the development and test sets is supported by examples in the train set. We believe this split strikes a good balance between challenge and feasibility: to succeed, a model needs to learn to identify relations in the complex sentence, link them to their arguments, and produce a rephrasing of them. However, it is not required to generalize to unseen relations. \footnote{The updated dataset (v1.0, published by Narayan et al. after this work was accepted) follows (2) above, but not (1).}

The data split and scripts for creating it are available on Github.\footnote{\url{https://github.com/biu-nlp/sprp-acl2018}} Statistics describing the data split are detailed in Table \ref{tab:RDF_split_stats}.

\begin{table}[t!]
\centering
{\footnotesize
\centering
\begin{tabular}{l|l|l}
                         & count & unique \\\hline
train complex sentences  & 1,039,392  & 4,506 \\%(81.28\%)
train simple sentences   & 5,239,279  & 7,865 \\
dev complex sentences    & 13,294     & 535   \\% (9.65\%)
dev simple sentences     & 39,703     & 812   \\ 
test complex sentences   & 13,429     & 503   \\ %(9.07\%)
test simple sentences    & 41,734     & 879   \\
\hline 
\# dev simple in train   &\multicolumn{2}{l}{ 35 (0.09\%)} \\
\# test simple in train  &\multicolumn{2}{l}{  1 (0\%)}  \\
\% dev vocab in train    &\multicolumn{2}{l}{  62.99\% } \\
\% test vocab in train    &\multicolumn{2}{l}{ 61.67\% } \\
\hline
dev entities in train    &\multicolumn{2}{l}{ 26/111 (23.42\%) } \\
test entities in train    &\multicolumn{2}{l}{ 25/120 (20.83\%) } \\
dev relations in train    &\multicolumn{2}{l}{ 34/34 (100\%) } \\
test relations in train    &\multicolumn{2}{l}{ 37/37 (100\%) } \\
\end{tabular}
}
\vspace{-0.2cm}
\caption{Statistics for the RDF-based data split }
\label{tab:RDF_split_stats}
\vspace{-0.6cm}
\end{table}

\section{Copy-augmented Model}\label{sec:copy}
To better suit the split-and-rephrase task, we augment the \seqseq{} models with a copy mechanism. Such mechanisms have proven to be beneficial in similar tasks like abstractive summarization \cite{gu2016incorporating,abc2017} and language modeling \cite{merity2016pointer}. 
We hypothesize that biasing the model towards copying will improve performance, as many of the words in the simple sentences (mostly corresponding to entities) appear in the complex sentence, as evident by the relatively high BLEU scores for the \textsc{Source} baseline in Table \ref{tab:results}. 

Copying is modeled using a ``copy switch'' probability $p(z)$ computed by a sigmoid over a learned composition of the decoder state, the context vector and the last output embedding. It interpolates the $p_{softmax}$ distribution over the target vocabulary and a copy distribution $p_{copy}$ over the source sentence tokens. $p_{copy}$ is simply the computed attention weights. Once the above distributions are computed, the final probability for an output word $w$ is:
\begin{align*}
p(w) = p(z=1) p_{copy}(w) + p(z=0) p_{softmax}(w)
\label{eq:pw}
\end{align*}
In case $w$ is not present in the output vocabulary, we set $p_{softmax}(w)=0$. We refer the reader to \newcite{abc2017} for a detailed discussion regarding the copy mechanism.

\section{Experiments and Results}
Models with larger capacities may have greater representation power, but also a stronger tendency to memorize the training data. We therefore perform experiments with copy-enhanced models of varying LSTM widths (128, 256 and 512).
We train the models using the negative log likelihood of $p(w)$ as the objective. Other than the copy mechanism, we keep the settings identical to those in Section \ref{sec:experimental_details}. 
We train models on the original split, our proposed data split and the v1.0 split.
% \footnote{This split was published after this paper was accepted. It is using a larger dataset than the original split and is also not allowing overlapping train/development/test RDFs, similarly to our new split.}.

% results with copy mechanism
\begin{table}[t!]
{\footnotesize
\begin{center}
\begin{tabular}{l l | c c c }
& &\bleu{} & \#S/C & \#T/S \\\cline{2-5}
\multirow{9}{*}{\rotatebox[origin=c]{90}{original data split}} 
& \textsc{Source}               & 55.67 & 1.0  & 21.11 \\
& \textsc{Reference}            & --    & 2.52 & 10.93 \\
& \splitseqseq                  & 78.77 & 2.84 & 9.28  \\ \cline{2-5}
& \textsc{Seq2Seq128} 	        & 76.56 & 2.53 & 10.53 \\ 
& \textsc{Seq2Seq256} 	        & 77.48 & 2.57 & 10.56 \\ 
& \textsc{Seq2Seq512} 	        & 75.92 & 2.59 & 10.59 \\ \cline{2-5}
& \textsc{Copy128}              & 78.55 & 2.51 & 10.29 \\ 
& \textsc{Copy256}              & 83.73 & 2.49 & 10.66 \\ 
& \textsc{Copy512} 		        & 87.45 & 2.56 & 10.50 \\ \hhline{~====} 
\multirow{8}{*}{\rotatebox[origin=c]{90}{new data split}} 
& \textsc{Source}               & 55.66 & 1.0  & 20.37  \\
& \textsc{Reference}            & --    & 2.40 & 10.83  \\ \cline{2-5}
& \textsc{Seq2Seq128}           & 5.55  & 2.27 & 11.68  \\ 
& \textsc{Seq2Seq256}           & 5.28  & 2.27 & 10.54  \\ 
& \textsc{Seq2Seq512}           & 6.68  & 2.44 & 10.23  \\ \cline{2-5}
& \textsc{Copy128}              & 16.71 & 2.0  & 10.53  \\ 
& \textsc{Copy256}              & 23.78 & 2.38 & 10.55  \\ 
& \textsc{Copy512} 	            & 24.97 & 2.87 & 10.04 \\ \hhline{~====}
\multirow{3}{*}{\rotatebox[origin=c]{90}{v1.0}} 
& \textsc{Source} 	            & 56.1  & 1.0  & 20.4  \\
& \textsc{Reference} 	        & --    & 2.48 & 10.69 \\ \cline{2-5}
& \textsc{Copy512} 	            & 25.47 & 2.29 & 11.74 \\
\end{tabular}
\end{center}
}
\vspace{-0.3cm}
\caption{
% Average \textsc{BLEU} scores, average number of sentences in the output texts (\#S/C) and average number of tokens per output sentences (\#T/S) 
Results over the test sets of the original, our proposed split and the v1.0 split}\label{tab:results_shuffled}
\vspace{-0.5cm}
\end{table}

% examples table
\begin{table*}[tbp]
\centering
{\footnotesize
\begin{tabular}{p{6.5cm}|p{8.5cm}}

% \hline
Input & Prediction \\
\hline
\vspace{-0.19cm}{\color{blue}Madeleine L'Engle who is influenced by George Macdonald wrote `` A Severed Wasp '' }.
&\begin{tabular}[t]{p{9cm}} \vspace{-0.2cm}{\color{red}A Severed Wasp was written by George Macdonald .}\\\vspace{-0.2cm}{\color{red}A Severed Wasp is from the United States .}
\end{tabular} \\
\hline
The A.C. Lumezzane has 4150 members and {\color{blue}play in the Lega Pro League} .&
\begin{tabular}[t]{p{9.3cm}}
\vspace{-0.26cm}{\color{red}A.C. Lumezzane 's ground is in the Lega Pro League .}\\
A.C. Lumezzane has 4150 members .\\
\end{tabular} \\ 
\hline
\vspace{-0.26cm}{\color{blue}Irish} English is the official language of Ireland , {\color{blue}which is lead by Enda Kenny} and home to Adare Manor .
& 
\begin{tabular}[t]{p{9.3cm}}
Adare Manor is located in Ireland . \\
English is the language of Ireland . \\
\vspace{-0.26cm}{\color{darkgreen}English is the language of Ireland . (repeated x3)} \\
% \vspace{-0.26cm}{\color{darkgreen}English is the language of Ireland .} \\
% \vspace{-0.26cm}{\color{darkgreen}English is the language of Ireland .}\\
\end{tabular} \\ 
\end{tabular}
}%end footnotesize
\vspace{-8px}
\caption{Predictions from the \textsc{Copy512} model, trained on the new data split.}  
\label{tab:issues_new_split}
\vspace{-15px}
\end{table*}
\paragraph{Results}
Table \ref{tab:results_shuffled} presents the results. On the original data-split, the \textsc{Copy512} model outperforms all baselines, improving over the previous best by 8.68 BLEU points. On the new data-split, as expected, the performance degrades for all models, as they are required to generalize to sentences not seen during training. The copy-augmented models perform better than the baselines in this case as well, with a larger relative gap which can be explained by the lower lexical overlap between the train and the test sets in the new split. On the v1.0 split the results are similar to those on our split, in spite of it being larger (1,331,515 vs. 886,857 examples), indicating that merely adding data will not solve the task.
% \vspace{-1cm}
\paragraph{Analysis}
We inspect the models' predictions for the first 20 complex sentences of the original and new validation sets in Table \ref{tab:analysis_copy}. We mark each simple sentence as being ``correct'' if it contains all and only relevant information, ``unsupported'' if it contains facts not present in the source, and ``repeated'' if it repeats information from a previous sentence. We also count missing facts. Figure \ref{fig:copy_mech_attn} shows the attention weights of the \textsc{Copy512} model for the same sentence in Figure \ref{fig:vanilla_attn}. Reassuringly, the attention is now distributed more evenly over the input symbols. 

On the new splits, all models perform catastrophically. Table \ref{tab:issues_new_split} shows outputs from the \textsc{Copy512} model when trained on the new split. On the original split, while \textsc{Seq2Seq128} mainly suffers from missing information, perhaps due to insufficient memorization capacity, \textsc{Seq2Seq512} generated the most unsupported sentences, due to overfitting or memorization. The overall number of issues is clearly reduced in the copy-augmented models.
% \vspace{-0.4cm}
% attention weights from the copy-augmented model
\InsertBoxC{
\begin{tabular}{c}\\
\includegraphics[scale=0.34]{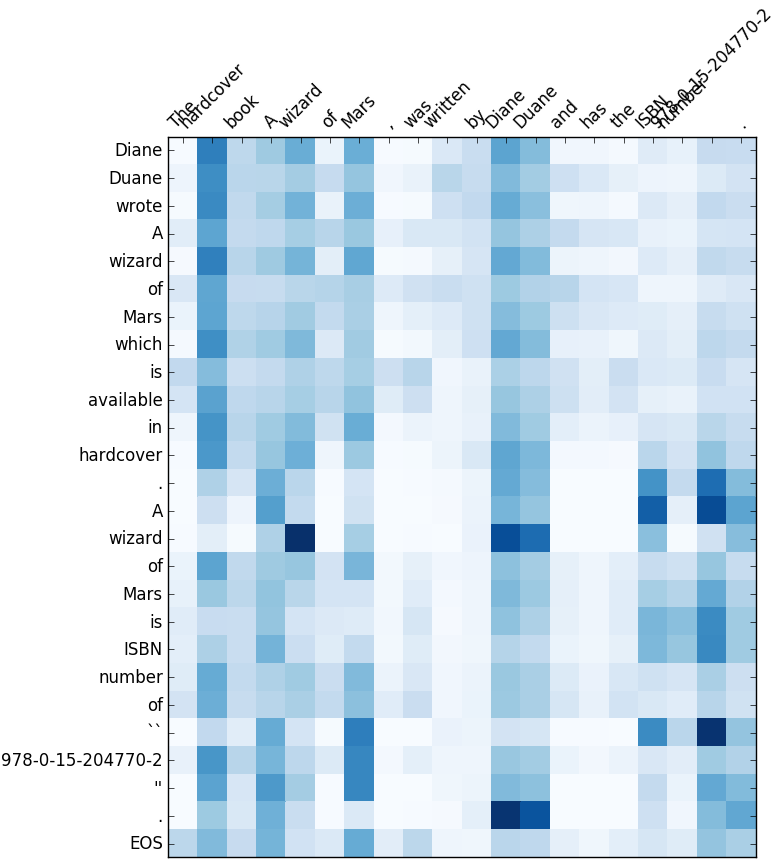}\\
\captionsetup{type=figure}
\parbox{\linewidth}{\captionof{figure}{Attention weights from the \textsc{Copy512} model for the same input as in Figure \ref{fig:vanilla_attn}.} 
% \vspace{-1.5cm}
\label{fig:copy_mech_attn}}\end{tabular}}
\vspace{-1.0cm}
\InsertBoxC{
\begin{tabular}{c}\\
\footnotesize
\begin{tabular}{P{1.8cm}|P{0.55cm} P{0.75cm} P{1.6cm} P{0.9cm}}
Model & unsup. & repeated & correct & missing \\\hline
\textbf{original split}      \\ 
\textsc{Seq2Seq128} & 5   & 4 & 40/49 (82\%) & 9   \\ 
\textsc{Seq2Seq256} & 2   & 2 & 42/46 (91\%) & 5   \\ 
\textsc{Seq2Seq512} & 12  & 2 & 36/49 (73\%) & 5   \\ 
\textsc{Copy128}    & 3   & 4 & 42/49 (86\%) & 4   \\
\textsc{Copy256}    & 3   & 2 & 45/50 (90\%) & 6   \\
\textsc{Copy512}    & 5   & 0 & 46/51 (90\%) & 3   \\\hline
\textbf{new split}       \\
\textsc{Seq2Seq128} & 37  & 8  & 0 & 54   \\
\textsc{Seq2Seq256} & 41  & 7  & 0 & 54   \\
\textsc{Seq2Seq512} & 43  & 5  & 0 & 54   \\
\textsc{Copy128}    & 23  & 3  & 2/27 (7\%)   & 52   \\
\textsc{Copy256}    & 35  & 2  & 3/40 (7\%)   & 49   \\
\textsc{Copy512}    & 36  & 13 & 11/54 (20\%) & 43   \\\hline
\textbf{v1.0 split}    		\\
\textsc{Copy512}    & 41  & 3  & 3/44 (7\%)   & 51   \\
\end{tabular}\\
\captionsetup{type=figure}
\parbox{\linewidth}{\vspace{5px}\captionof{table}{Results of the manual analysis, showing the number of simple sentences with unsupported facts (unsup.), repeated facts, missing facts and correct facts, for 20 complex sentences from the original and new validation sets.} 
\vspace{-15px}
\label{tab:analysis_copy}}\end{tabular}}
\section{Conclusions}
\vspace{-0.2cm}
We demonstrated that a \seqseq{} model can obtain high scores on the original split-and-rephrase task while not actually learning to split-and-rephrase. We propose a new and more challenging data-split to remedy this, and demonstrate that the cheating \seqseq{} models fail miserably on the new split. Augmenting the \seqseq{} models with a copy-mechanism improves performance on both data splits, establishing a new competitive baseline for the task. Yet, the split-and-rephrase task (on the new split) is still far from being solved. We strongly encourage future research to evaluate on our proposed split or on the recently released version 1.0 of the dataset, which is larger and also addresses the overlap issues mentioned here.

\subsubsection*{Acknowledgments}
We thank Shashi Narayan and Jan Botha for their useful comments. The work was supported by the Intel Collaborative Research Institute for Computational Intelligence (ICRI-CI), the Israeli Science Foundation (grant number 1555/15), and the German Research Foundation via the German-Israeli Project Cooperation (DIP, grant DA 1600/1-1).

%Future work may explore sub-word \cite{sennrichbpe} and character level \cite{lee2016fully} modeling to improve generalization to unseen words, or incorporate syntactic information \cite{bastings2017graph,aharoni-goldberg:2017:Short}. 
% In addition, further improvements can be made in the dataset to include more real-world complex sentences to increase its linguistic diversity. 
\bibliography{sprp}
\bibliographystyle{acl_natbib}

\newpage
\section*{Appendix A}

\section*{Training details}
Outr models are trained with early stopping by running the proposed evaluation method on the development set after every epoch. We use a single layer LSTM for the encoder and decoder. We tie the embeddings of the encoder and the decoder, and preliminary experiments showed similar results without tying. In all models The size of the embedding vectors is similar to the size of the LSTM units (128/256/512). We decode using beam search with a beam size of 12. All model parameters, including the embeddings are randomly initialized and learned during training. For optimization we use SGD with an initial learning rate of 1.0 and decay the learning rate by 0.5 when there is no improvement on the validation set.

\end{document}